\documentclass[11pt]{article}

\usepackage[preprint]{acl}

\usepackage{times}
\usepackage{latexsym}

\usepackage[T1]{fontenc}
\usepackage[utf8]{inputenc}

\usepackage{microtype}

\usepackage{graphicx}
\usepackage{algorithm}
\usepackage{algorithmic}
\usepackage{amsmath}
\usepackage{amssymb}
\usepackage{xcolor}
\usepackage{subcaption}
\usepackage{multirow}
\usepackage{booktabs}
\usepackage{placeins}

\title{PACT: Learning Diverse Diagnostic Strategies via Privileged Synthesis and Branch Consensus}

\author{Gen Li$^1$\quad
  Yuanze Hu$^1$\quad
  Zhichao Yang$^1$\quad
  Qingchen Yu$^1$\quad
  Jianwei Lv$^2$\quad
  Yue Guo$^2$\quad
  Yujing Liu$^3$\\
  Faguo Wu$^1$\quad
  Hongwei Zheng$^4$\quad
  Xiandong Li$^{2\dagger}$\quad
  Bo Yuan$^{2\dagger}$\quad
  Yifan Sun$^{5*}$\quad
  Zhaoxin Fan$^{1*}$\\[6pt]
  $^1$Beihang University\quad
  $^2$Baidu\quad
  $^3$ByteDance\\
  $^4$Beijing Academy of Blockchain and Edge Computing\quad
  $^5$Renmin University of China\\[4pt]
  {\small $^\dagger$Project leaders.\quad $^*$Corresponding authors.}\\
  {\small Correspondence: \texttt{sunyifan1984@163.com}, \texttt{zhaoxinf@buaa.edu.cn}}
}

\begin{document}

\maketitle
\begin{abstract}
Clinical diagnosis requires flexible use of multiple reasoning paradigms under incomplete patient information. Existing LLM-based medical agents show strong medical reasoning ability, but single-paradigm or naively mixed dialogue supervision makes these paradigms difficult to learn without interference. We propose \textbf{PACT} (Periodic Anchor Consensus Training), a framework that couples supervised multi-paradigm dialogue synthesis with consensus-based Branch training. At the data level, \textbf{DPS} (Doctor-Patient-Supervisor) uses complete electronic medical records (EMRs) for quality control while keeping the doctor agent restricted to patient-visible information. This produces validated dialogues under four diagnostic reasoning paradigms without leaking hidden clinical answers. At the training level, PACT trains one paradigm-specific LoRA Branch per paradigm and periodically aggregates Branches into a shared Anchor through sign consensus. We further construct a dynamic multi-turn Chinese medical diagnosis benchmark for interactive consultation. Experiments show that PACT achieves state-of-the-art performance among compared proprietary, medical-specialized, and task-adapted baselines on diagnostic outcome and consultation-process metrics.
\end{abstract}

\begin{figure}[!t]
    \centering
    % Top: trend line
    \begin{subfigure}[b]{0.9\columnwidth}
        \includegraphics[width=\linewidth]{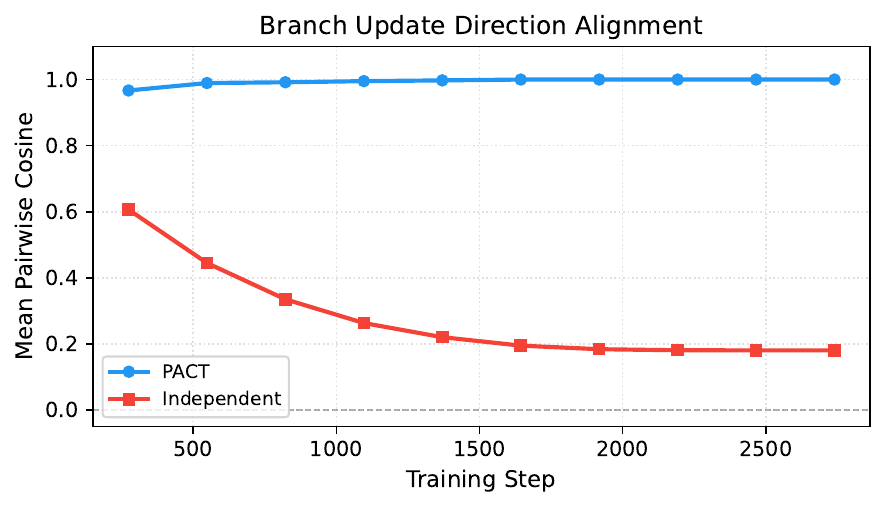}
        \vspace{-8mm}
        \caption{Mean pairwise cosine across training.}
        \label{fig:teaser_a}
    \end{subfigure}

    % \vspace{-4mm}
    % Bottom: compact heatmap grid
    \begin{subfigure}[b]{0.95\columnwidth}
        \includegraphics[width=\linewidth]{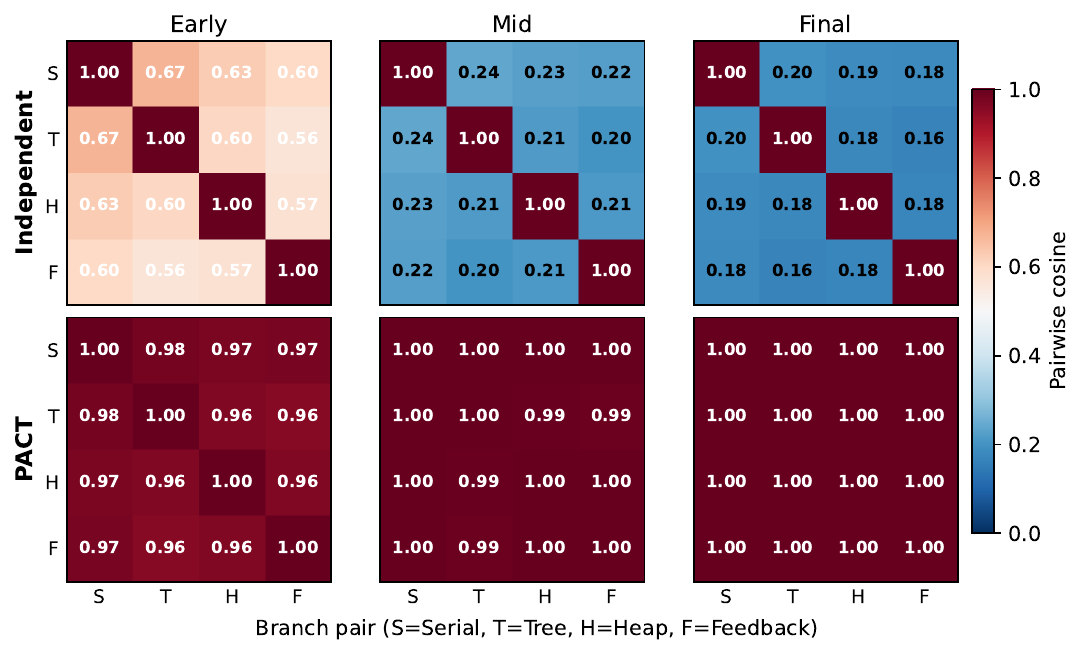}
        \caption{Checkpoint heatmaps.}
        \label{fig:teaser_b}
    \end{subfigure}
    \caption{Branch alignment during training. Each Branch corresponds to one diagnostic reasoning mode. Independent training causes Branch updates to diverge, whereas PACT maintains high alignment and merge-compatible update directions.}
    % \vspace{-4mm}
    \label{fig:teaser}
\end{figure}

\section{Introduction}

The rapid expansion of online medical consultation platforms has created growing demand for diagnostic dialogue systems that can conduct safe and effective multi-turn consultations \cite{zeng2020meddialog,valizadeh2022survey}. Unlike single-turn medical question answering, clinical diagnosis is a conversation under uncertainty: the doctor must ask follow-up questions, collect missing evidence, and decide when enough information has been gathered \citep{croskerry2009universal,norman2005research}. Although large language models (LLMs) have shown promising medical reasoning abilities \cite{zhang2023huatuogpt,achiam2023gpt4}, strong medical knowledge does not necessarily imply knowing what to ask next in multi-turn consultation \citep{guo2026drassistant,lai2026doctorr1}. Learning multiple diagnostic reasoning paradigms in one deployable model therefore remains a key challenge.

A key reason is that clinical diagnosis involves multiple reasoning paradigms rather than a single template. A doctor may proceed step by step from symptoms to diagnosis, compare several possible diseases in parallel, ask the most urgent question first, or revise earlier hypotheses after receiving new evidence \cite{croskerry2009universal,norman2005research}. An LLM-based doctor should therefore learn these reasoning paradigms and draw on them during consultation. This central challenge decomposes into two sub-problems: how to synthesize training dialogues that demonstrate different reasoning paradigms, and how to train one model to learn these paradigms without making them interfere with each other?

The challenge in data synthesis lies in obtaining controlled diagnostic trajectories without violating doctor-patient information asymmetry. Clinician-authored diagnostic dialogues are expensive and scarce, and existing medical dialogue datasets are often collected from natural online consultations rather than controlled diagnostic processes \cite{zeng2020meddialog,liu2022meddg}. LLM-based synthesis is a practical alternative, but it has a basic information-asymmetry problem. To generate a medically correct dialogue, the synthesis system needs access to the complete electronic medical record (EMR), including the final diagnosis and examination results. However, the simulated doctor in the dialogue should not see this hidden information; otherwise, it may ask questions that implicitly leak the answer. In contrast, if the doctor and patient agents freely role-play without supervision, the dialogue may become repetitive, hallucinated, or clinically invalid.

The challenge in training lies in integrating multiple reasoning paradigms into a single deployable model. Even with multi-paradigm dialogues, simply mixing them for supervised fine-tuning can blur paradigm-specific learning. Training separate Branches---one LoRA adapter per reasoning paradigm---preserves specialization, but independently trained Branches may move in incompatible directions and become hard to combine into one deployable model. Sequentially training on one paradigm after another can also bias the model toward later paradigms. Figure~\ref{fig:teaser} shows this conflict empirically: under independent training, Branch update directions diverge to a mean pairwise cosine of 0.18, and 38\% of parameters conflict during TIES merging. Thus, reasoning paradigms require training-time periodic aggregation rather than only data-level collection.

To address these problems, we propose \textbf{PACT} (Periodic Anchor Consensus Training), a framework that couples supervised multi-paradigm dialogue synthesis with consensus-based Branch training. At the data level, \textbf{DPS} (Doctor-Patient-Supervisor) addresses the synthesis problem through role-separated supervision inspired by Learning Using Privileged Information (LUPI) \cite{vapnik2009lupi}, preserving doctor-patient information asymmetry while yielding validated dialogues under four reasoning paradigms---\textit{Serial}, \textit{Tree}, \textit{Heap}, and \textit{Feedback}.

At the training level, PACT trains one LoRA Branch for each reasoning paradigm. These Branches are periodically aggregated into a shared Anchor through sign-consensus merging, and an L1 regularization term keeps Branches close to the Anchor between aggregation rounds. The final Anchor is deployed as a standard single-LoRA model, requiring no routing mechanism, architectural change, or additional inference overhead.

Existing medical benchmarks mainly test single-turn QA or static diagnosis, leaving limited support for evaluating whether an agent can collect missing evidence through multi-turn consultation. We therefore introduce a dynamic multi-turn Chinese medical diagnosis benchmark to evaluate diagnostic agents under interactive consultation. Our contributions are:

\begin{itemize}
    \item We propose DPS, a distillation-based data-construction module that synthesizes validated dialogues under controlled reasoning paradigms while preserving doctor-patient information asymmetry.
    \item We propose PACT, a consensus-based LoRA training method that learns paradigm-specific Branches and periodically aggregates them into one deployable Anchor without extra inference cost.
    \item We introduce a dynamic multi-turn Chinese medical diagnosis benchmark and compare task-specific adaptation with zero-shot proprietary and specialized baselines under the same simulator-based protocol.
\end{itemize}

\section{Related Work}\label{sec:related_work}

\paragraph{Medical dialogue data and synthetic supervision.}
Large-scale healthcare dialogue corpora such as MedDialog \cite{zeng2020meddialog} and MedDG \cite{liu2022meddg} lack controlled diagnostic trajectories or preservation of doctor-patient information asymmetry. Recent medical LLMs improve instruction following and reasoning \cite{zhang2023huatuogpt,chen2023huatuogpt2,chen2024huatuogpto1,zhang2024ultramedical,jeong2024selfbiorag}, yet their supervision remains closer to single-turn QA than interactive diagnosis. LLM-based synthesis can scale clinician-level supervision, but single-generator pipelines risk leaking privileged labels, while unconstrained role-play drifts into clinically invalid trajectories. Dr.\ Assistant \citep{guo2026drassistant} shows that guideline-distilled reasoning data with RL improves multi-turn inquiry. In contrast, DPS constructs EMR-grounded validated dialogues under multiple reasoning paradigms while preserving doctor-patient information asymmetry through LUPI-style privileged supervision \cite{vapnik2009lupi} and lightweight LLM-as-a-Judge intervention \cite{zheng2023judgellm}.

\paragraph{Clinical reasoning strategies.}
Clinical diagnosis has long been viewed as a heterogeneous reasoning process involving hypothesis testing, differential diagnosis, test-threshold decisions, and reflection \cite{elstein1978medical,bowen2006clinical,pauker1980threshold,schon1983reflective,croskerry2009universal,norman2005research}. Prompting methods such as chain-of-thought and tree-of-thought operationalize related ideas at inference time \cite{wei2022chain,yao2023tree}. Our four reasoning paradigms---Serial, Tree, Heap, and Feedback---move these complementary diagnostic modes to the training-data level, exposing the model to paradigm-specific supervision rather than a single undifferentiated mixture.

\paragraph{Parameter-efficient adaptation and branch aggregation.}
LoRA \cite{hu2022lora} enables training multiple task-specific adapters, but combining them into single model remains difficult. Post-hoc merging methods like model soups, task arithmetic, SLERP, DARE, and TIES \cite{wortsman2022soups,ilharco2023task,goddard2024arcee,yu2024dare,yadav2023ties} assume that independently trained models are already merge-compatible. Related work in multi-task and federated optimization studies gradient conflict and client drift \cite{yu2020pcgrad,liu2021cagrad,mcmahan2017fedavg,li2020fedprox,karimireddy2020scaffold}. PACT instead treats adapter compatibility as a training-time objective, replacing one-shot post-hoc merging with periodic aggregation over paradigm-specific LoRA Branches.
\begin{figure*}[!htbp]
    \centering
    \includegraphics[width=\linewidth]{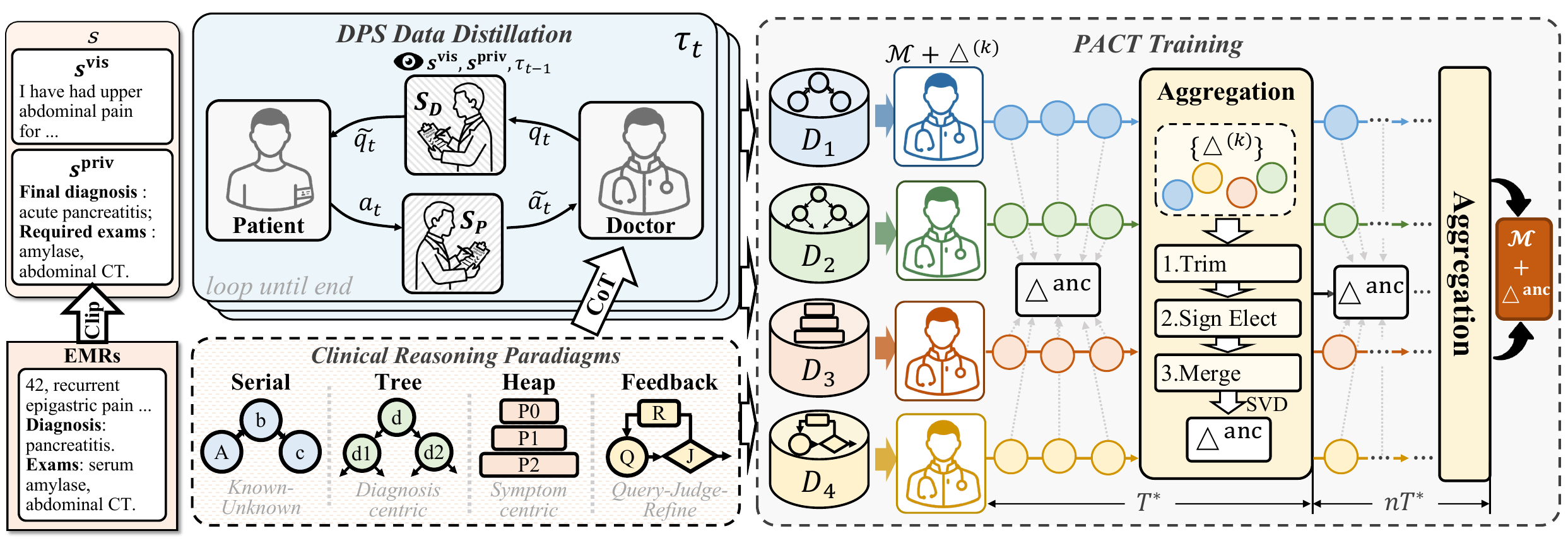}
    \caption{Overview of the PACT framework. \textbf{Left:} DPS splits each EMR into patient-visible memory and privileged clinical state, generates paradigm-conditioned Doctor--Patient dialogues, and uses Patient/Doctor Supervisors for minimal PASS-or-REWRITE quality control. \textbf{Right:} PACT trains one Branch LoRA per reasoning paradigm and periodically aggregates Branches into a global Anchor via sign-consensus merging, with L1 regularization constraining Branch drift between aggregation rounds.}
    \label{fig:pipeline}
    \vspace{-2mm}
\end{figure*}
\section{Methodology}\label{sec:methodology}

In this section, we first formulate multi-turn diagnosis as a partially observable decision process. Then we introduce DPS, which constructs information-asymmetric validated dialogues under four reasoning paradigms, and PACT, which trains one Branch per reasoning paradigm and periodically aggregates Branches into a shared Anchor for single-LoRA deployment. Figure~\ref{fig:pipeline} gives an overview of the full DPS--PACT pipeline.

\subsection{Problem Formulation}\label{sec:task_formulation}

Given a hidden clinical state and an observable dialogue history, our goal is to train a diagnostic agent that asks informative follow-up questions and produces an accurate diagnosis within a limited number of turns. We formalize this as a partially observable multi-turn decision process.

Each case has a hidden clinical state $s$ that includes the complete medical record, examination results, and target diagnosis $d^*$. The diagnostic agent cannot observe $s$ directly. A consultation begins with the patient's chief complaint $c_0$; at each turn $t$, the doctor issues a question $q_t$ and receives a patient answer $a_t$, building the trajectory
\begin{equation}
\label{eq:trajectory}
\tau_t = (c_0,\; q_1, a_1,\; \ldots,\; q_t, a_t).
\end{equation}
After at most $T_{\max}$ turns, the doctor terminates with a diagnosis $\hat{d}$ and examination recommendation $\hat{e}$. Let $\pi$ denote the diagnostic policy that maps the current trajectory to the next doctor action, including follow-up questioning, stopping, diagnosis, and examination recommendation. The objective is
\begin{equation}
\label{eq:objective}
\max_\pi\; P(\hat{d}=d^* \mid \tau_T;\, \pi), \quad T \leq T_{\max}.
\end{equation}
This formulation highlights the two requirements addressed below: reliable data must preserve the information gap between $s$ and $\tau_t$, and training must integrate different reasoning paradigms into one model. Next, we introduce DPS for constructing information-asymmetric paradigm-specific dialogues, followed by PACT for training a single deployable model from these dialogues.

\subsection{DPS Data Synthesis Architecture}\label{sec:dps_architecture}

DPS is designed to synthesize validated dialogues without exposing hidden EMR information to the simulated Doctor. The key tension is that dialogue generation needs the complete clinical trajectory for quality control, but the Doctor role should only see patient-visible information. DPS resolves this tension through visible/privileged state separation, role-separated dialogue generation, and minimal supervisor validation. The resulting dialogues are conditioned on four reasoning paradigms and later used as PACT training data.

\textbf{Visible and privileged state.} To enforce this information boundary, DPS first partitions each EMR into patient-visible information $s^{\text{vis}}$ (chief complaint, symptom history, physical signs) and privileged clinical information $s^{\text{priv}}$ (unrevealed examinations, target diagnosis, treatment plan):
\begin{equation}
\label{eq:state_partition}
s = \bigl(s^{\text{vis}},\; s^{\text{priv}}\bigr).
\end{equation}

\textbf{Role-separated dialogue generation.} Given this partition, DPS then generates the dialogue with separated Doctor and Patient roles. At turn $t$, the Doctor drafts a question $q_t$ from the accumulated trajectory $\tau_{t-1}$, and the Patient drafts an answer $a_t$ from $s^{\text{vis}}$ only. Thus, the Doctor never observes $s^{\text{priv}}$. Let $S_D$ and $S_P$ denote the Doctor-Supervisor and Patient-Supervisor, respectively; DPS applies them as minimal PASS-or-REWRITE mappings:
\begin{equation}
\label{eq:dps_supervision_q}
\tilde{q}_t = S_D(q_t;\tau_{t-1},s),
\end{equation}
\begin{equation}
\label{eq:dps_supervision_a}
\tilde{a}_t = S_P(a_t;s^{\text{vis}},\tilde{q}_t).
\end{equation}

\textbf{Supervisor validation.} After each draft exchange, the Supervisors validate whether the Doctor question and Patient answer should be kept. If an output passes inspection, it is unchanged; otherwise, the Supervisor makes a local correction such as removing leakage, adjusting pacing, or adding a missing high-value question, without rewriting the whole response or injecting unrevealed patient facts. The validated exchange is then appended to the observable trajectory:
\begin{equation}
\label{eq:dps_trajectory_update}
\tau_t = (\tau_{t-1},\tilde{q}_t,\tilde{a}_t).
\end{equation}

Privileged information therefore affects the dialogue only through non-conversational process supervision. This design reduces direct label leakage, though privileged supervision can still introduce trajectory-level bias; we analyze this in Appendix~\ref{sec:failure_modes}, and DPS data quality in Table~\ref{tab:dps_quality}.

\textbf{Reasoning-paradigm conditioning.} Finally, DPS repeats the same information-asymmetric generation under four reasoning paradigms inspired by clinical reasoning theories \cite{elstein1978medical,bowen2006clinical,pauker1980threshold,schon1983reflective}: \textit{Serial}, \textit{Tree}, \textit{Heap}, and \textit{Feedback}. Serial follows stepwise hypothesis testing; Tree maintains competing diagnoses for differential inquiry; Heap prioritizes questions by urgency and expected information value; Feedback performs a brief self-critique before responding. We do not assume that the generated dialogues are perfectly separable by paradigm; instead, the paradigms serve as controlled sources of variation for training. Prompt templates and generation controls are summarized in Appendix~\ref{sec:prompt_templates}. We denote the validated dialogues collected under paradigm $k$ as $D_k$; the $K{=}4$ datasets $\{D_k\}_{k=1}^K$ are passed to PACT as training inputs.

\subsection{PACT: Periodic Anchor Consensus Training}\label{sec:pact}

\begin{figure}[t]
    \centering
    \includegraphics[width=0.95\columnwidth]{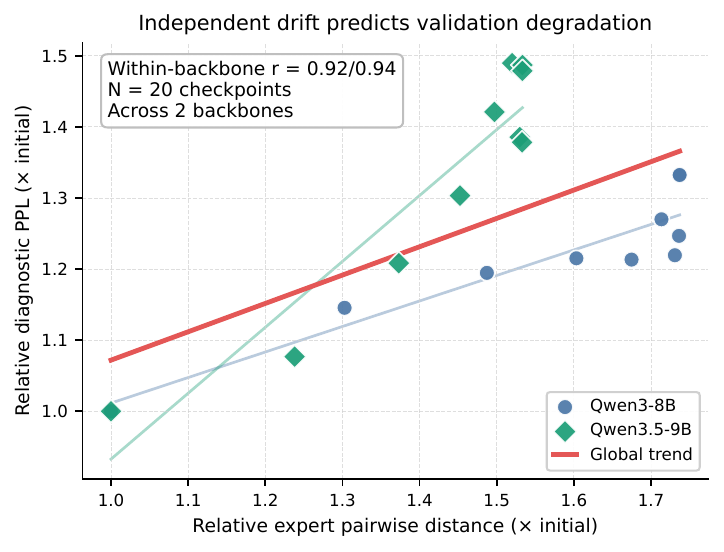}
    \caption{Pilot study on independent fine-tuning. Across Qwen3-8B and Qwen3.5-9B, increasing Branch pairwise distance correlates with diagnostic validation PPL degradation. Distances and PPL are normalized by the first checkpoint within each backbone for cross-model comparability.}
    \label{fig:pilot}
    \vspace{-2mm}
\end{figure}

After DPS constructs the $K{=}4$ paradigm-specific datasets $\{D_k\}_{k=1}^K$, the next challenge is to train one deployable model from them. Two direct strategies are Mixed-SFT, which trains one adapter on the union of all dialogues, and independent Branch training, which trains one adapter per reasoning paradigm. The former can blur paradigm-specific learning, while the latter can produce Branches that drift into incompatible update directions. PACT therefore alternates between local Branch training with Anchor regularization and periodic Anchor aggregation through sign-consensus merging. Figure~\ref{fig:pilot} shows that larger Branch distance correlates with worse diagnostic validation PPL; parameter-space drift trends are further analyzed in Appendix Figure~\ref{fig:pca_comparison}.

\textbf{Branch training with Anchor regularization.} We first train one LoRA Branch (a paradigm-specific adapter) $\Delta^{(k)}$ for each reasoning paradigm while maintaining a shared Anchor $\Delta^{\text{anc}}$ as the deployable consensus adapter. Both are initialized to zero at the start of training, so the initial L1 penalty regularizes Branches toward the base model. Each Branch minimizes:
\begin{equation}
\label{eq:total_loss}
\mathcal{L}_{\text{total}}^{(k)} = \mathcal{L}_{\text{task}}^{(k)} + \lambda \cdot \| \Delta^{(k)} - \Delta^{\text{anc}} \|_1,
\end{equation}
where $\mathcal{L}_{\text{task}}^{(k)}$ is the next-token cross-entropy loss on $D_k$, and $\lambda$ controls consensus strength. The L1 penalty encourages sparse paradigm-specific deviations while keeping most Branch directions close to the Anchor.

\textbf{Periodic sign-consensus aggregation.} After local Branch updates, PACT periodically refreshes the Anchor. Every $T^*$ epochs, it updates the Anchor from the current Branch deltas through three sign-consensus steps: trim, elect, and merge-project.

\textbf{Step 1: Trim.} The first step removes low-magnitude LoRA updates that mainly reflect optimization noise rather than paradigm-specific learning. Retaining only the top-$\rho$ fraction of updates by magnitude reduces noise interference before aggregation:
\begin{equation}
\label{eq:trim}
\hat{\Delta}^{(k)} = \operatorname{Trim}(\Delta^{(k)}, \rho),
\end{equation}
where $\rho$ is a shared density hyperparameter applied to each Branch's own magnitude distribution independently.

\textbf{Step 2: Elect.} Next, PACT resolves conflicting update directions across Branches by majority vote on the sign of each parameter dimension:
\begin{equation}
\label{eq:sign}
\gamma_i = \operatorname{sign}\!\Bigl(\textstyle\sum_{k=1}^K \hat{\Delta}^{(k)}_i\Bigr).
\end{equation}
Only updates whose sign agrees with $\gamma_i$ contribute to aggregation, discarding contributions that would cause destructive interference.

\textbf{Step 3: Merge-project.} Finally, PACT averages only sign-consistent values and projects the result back to LoRA rank. Let $\mathcal{S}_i = \{k : \operatorname{sign}(\hat{\Delta}^{(k)}_i) = \gamma_i\}$ denote the set of sign-consistent Branches at position $i$. We first build a full consensus matrix $\bar{\Delta}$ by averaging only sign-consistent values at each coordinate:
\begin{equation}
\label{eq:aggregate_coord}
\bar{\Delta}_i = \frac{1}{|\mathcal{S}_i|}\sum_{k \in \mathcal{S}_i} \hat{\Delta}^{(k)}_i.
\end{equation}
The consensus matrix is then projected back to LoRA rank $r$ via truncated SVD:
\begin{equation}
\label{eq:aggregate}
\Delta^{\text{anc}} \leftarrow \operatorname{SVD}_r\!\bigl(\bar{\Delta}\bigr).
\end{equation}
The $\operatorname{SVD}_r$ projection is required because the coordinate-wise consensus matrix is generally full-rank; truncated SVD recovers the closest rank-$r$ approximation, maintaining valid LoRA structure. Branches are \emph{not} reset to the Anchor after aggregation: retaining local parameters preserves optimization continuity, while the updated Anchor redirects subsequent training toward the new consensus. Together, Anchor regularization controls local Branch drift, while periodic aggregation refreshes the deployable Anchor as training progresses. Algorithm~\ref{alg:pact} summarizes the full training loop.

\begin{algorithm}[t]
\caption{PACT: Periodic Anchor Consensus Training}
\label{alg:pact}
\begin{algorithmic}[1]
\REQUIRE Base model $\mathcal{M}$, datasets $\{D_k\}_{k=1}^{K}$, $\lambda$, $T^*$, TIES density $\rho$
\STATE Initialize Branches $\{\Delta^{(k)}\}_{k=1}^{K}$ and Anchor $\Delta^{\text{anc}}$
\FOR{each training phase}
    \STATE Train each Branch on $D_k$ with $\mathcal{L}_{\text{task}}^{(k)} + \lambda \|\Delta^{(k)} - \Delta^{\text{anc}}\|_1$
    \IF{phase reaches aggregation interval $T^*$}
        \STATE $\hat{\Delta}^{(k)} \leftarrow \operatorname{Trim}(\Delta^{(k)}, \rho)$ for each $k$ \hfill
        \STATE $\gamma_i \leftarrow \operatorname{sign}\!\bigl(\sum_k \hat{\Delta}^{(k)}_i\bigr)$ for each $i$ \hfill
        \STATE Apply sign-consensus merge and rank-$r$ projection to update $\Delta^{\text{anc}}$ \hfill
        \STATE Update the Anchor reference; Branches continue from their local parameters
    \ENDIF
\ENDFOR
\RETURN $\mathcal{M} + \Delta^{\text{anc}}$
\end{algorithmic}
\end{algorithm}

\textbf{Deployment.} After training, PACT discards all Branch adapters and deploys only $\mathcal{M} + \Delta^{\text{anc}}$. The deployed model is therefore identical to a standard single-LoRA model in architecture and inference cost, while the training process exposes the Anchor to all four reasoning paradigms through periodic aggregation. A detailed discussion is provided in Appendix~\ref{sec:theory_appendix}, implementation details are in Appendix~\ref{sec:pact_impl}, and checkpoint-level sign-conflict analysis is provided in Appendix~\ref{sec:sign_conflict}.

\section{Experiments}\label{sec:experiments}
\begin{table*}[!t]
    \centering
    \resizebox{\textwidth}{!}{
    \begin{tabular}{lc ccc ccc cc}
        \toprule
        \multirow{2}{*}{Method} & \multirow{2}{*}{Size} & \multicolumn{3}{c}{Diagnosis} & \multicolumn{3}{c}{Examination} & \multicolumn{2}{c}{Process} \\
        \cmidrule(lr){3-5} \cmidrule(lr){6-8} \cmidrule(lr){9-10}
        & & Strict DA & Relaxed DA & Soft Score & Exam F1 & Exam P & Exam R & Avg Rnd & Symp Cov \\
        \midrule
        \multicolumn{10}{l}{\textit{Proprietary LLMs}} \\
        Gemini-3.1 Pro    & --  & 29.00 & 63.93 & 52.63 & 9.22  & 11.03 & 8.74  & 6.61 & 83.50 \\
        DeepSeek-R1       & --  & 29.39 & 72.20 & 56.85 & 19.26 & 20.21 & 22.45 & 3.32 & 87.00 \\
        Doubao-Seed-1.6   & --  & 30.08 & 71.69 & 57.75 & \underline{25.71} & \underline{29.53} & 27.45 & 3.96 & 89.60 \\
        GPT-5             & --  & 33.45 & 64.10 & 54.06 & 12.83 & 14.45 & 13.92 & 5.64 & 79.02 \\
        GPT-4.1           & --  & \underline{36.70} & \underline{78.82} & \underline{64.86} & 17.26 & 20.40 & 19.76 & 3.85 & \textbf{94.04} \\
        \midrule
        \multicolumn{10}{l}{\textit{Medical-specialized LLMs}} \\
        Reflect-PubMed    & 8B  & 14.69 & 51.20 & 39.40 & 10.39 & 2.57  & 10.73 & 2.71 & 67.70 \\
        DoctorAgent-RL    & 8B  & 20.36 & 61.43 & 48.74 & 8.25  & 2.01  & 7.53  & 4.36 & 65.65 \\
        UltraMedical      & 8B  & 21.77 & 77.71 & 58.67 & 25.70 & 27.56 & \underline{28.07} & 3.73 & 71.27 \\
        HuatuoGPT2-7B     & 7B  & 27.85 & 63.64 & 51.24 & 9.99  & 11.71 & 10.47 & 1.80 & 54.39 \\
        HuatuoGPT-o1-8B   & 8B  & 30.54 & 70.61 & 57.26 & 3.21  & 1.54  & 3.36  & 1.72 & 63.88 \\
        Doctor-R1         & 8B  & 33.50 & 72.09 & 57.80 & 21.00 & 22.57 & 23.92 & 4.88 & 77.00 \\
        \midrule
        PACT & 8B  & \textbf{55.31}$^{\dagger}$ & \textbf{85.79}$^{\dagger}$ & \textbf{76.16}$^{\dagger}$ & \textbf{61.33}$^{\dagger}$ & \textbf{57.50} & \textbf{74.27} & 4.02 & \underline{93.39} \\
        \bottomrule
    \end{tabular}
    }
    \caption{Main results on dynamic multi-turn diagnostic evaluation. Best results are \textbf{bolded}, second best are \underline{underlined}. PACT is trained on Qwen3-8B. $^\dagger$ indicates significant improvement over GPT-4.1 under paired bootstrap ($p<0.05$).}
    \label{tab:main}
    \vspace{-2mm}
\end{table*}
\subsection{Experimental Setup}

Existing medical benchmarks mainly test single-turn QA or static diagnosis, and therefore do not measure whether a diagnostic agent can collect missing evidence through multi-turn consultation. We therefore construct a dynamic multi-turn Chinese diagnosis benchmark and evaluate all models under the same simulator-based protocol.

\textbf{Data.} We collect 10,000 de-identified Internal Medicine electronic medical records (EMRs). From this pool, we sample 6,357 EMRs for DPS synthesis, producing 25,428 validated dialogue candidates (4 paradigms per EMR). After quality filtering---removing EMRs where $\geq 2$ paradigm-specific dialogues fail diagnosis matching---we retain 20,684 validated dialogues from 5,171 unique EMRs, evenly distributed across the four reasoning paradigms. These 5,171 EMRs constitute the synthesis-derived training pool and are split at the EMR level into train/validation/test partitions with an 85\%/5\%/10\% ratio. The dynamic evaluation set is constructed separately from the remaining de-identified Internal Medicine pool and contains 438 unique held-out EMR cases; no EMR in this set appears in the DPS synthesis or training partitions. The retained synthesis pool covers 145 unique diagnostic labels.

\textbf{Model and Training.} We use Qwen3-8B \cite{qwen2025qwen3} as the primary base model, with Qwen3.5-9B, Llama3.1-8B, and Ministral-3-8B for cross-architecture analysis. LoRA adapters are configured with rank $r = 64$, $\alpha = 64$, and applied to all attention projection matrices. Training uses AdamW with learning rate $5 \times 10^{-4}$, cosine schedule, warmup ratio 0.03, and bfloat16 precision. The effective batch size is 64 (per-device 2 $\times$ gradient accumulation 8 $\times$ 4 GPUs). For PACT, we set the synchronization strength $\lambda = 200$, aggregation interval $T^* = 0.25$ epochs, and TIES density $\rho = 0.5$. Additional dataset, preprocessing, evaluation, and hyperparameter details are provided in Appendix~\ref{sec:data_eval_details}, with sensitivity to $\lambda$ and $T^*$ reported in Appendix Figure~\ref{fig:hyperparam}.

\textbf{Evaluation Protocol.} We adopt a \textit{dynamic multi-turn evaluation} protocol where the trained model interacts with a GPT-4o-powered patient simulator \cite{openai2024gpt4o} for up to 8 rounds per held-out EMR case. A separate GPT-4o judge scores the completed dialogue on three dimensions: \textit{Diagnosis} (Strict DA for exact top-1 match, Relaxed DA for hierarchical/top-3 credit, and Soft Score as a graded composite), \textit{Examinations} (Exam F1/Precision/Recall against the gold-standard list), and \textit{Process} (Symptom Coverage of key clinical information and average dialogue rounds). Detailed scoring rubrics are in Appendix~\ref{sec:data_eval_details}. Each EMR case is evaluated over four independent rollouts to reduce simulator variance; statistical tests (Appendix~\ref{sec:significance_tests}) are paired at the EMR-case level.

\textbf{Baselines.} We compare against \textit{proprietary LLMs} evaluated zero-shot under the same protocol: GPT-5 \cite{openai2025gpt5}, GPT-4.1 \cite{openai2025gpt41}, Gemini-3.1 Pro \cite{google2025gemini3}, DeepSeek-R1 \cite{guo2025deepseekr1}, and Doubao-Seed-1.6 \cite{bytedance2025doubao}; and \textit{medical-specialized LLMs}: HuatuoGPT2-7B \cite{chen2023huatuogpt2}, HuatuoGPT-o1-8B \cite{chen2024huatuogpto1}, UltraMedical \cite{zhang2024ultramedical}, Doctor-R1 \cite{lai2026doctorr1}, DoctorAgent-RL \cite{feng2025doctoragentrl}, and Reflect-PubMed \cite{jeong2024selfbiorag}. Cross-architecture results on general-purpose LLMs (Qwen3-8B, Qwen3.5-9B \cite{qwen2025qwen3}, Llama3.1-8B \cite{dubey2024llama3}, Ministral-3-8B \cite{liu2025ministral3}) are in Table~\ref{tab:cross_arch}.

\subsection{Main Results}

Table~\ref{tab:main} presents the main comparison across 11 baselines spanning proprietary and medical-specialized LLMs. PACT achieves 55.31\% Strict DA and 76.16 Soft Score, significantly exceeding the strongest zero-shot proprietary reference GPT-4.1 on Strict DA (+18.61, 95\% CI [15.24, 22.03], $p<0.0001$), Soft Score (+11.30, 95\% CI [8.68, 13.94], $p<0.0001$), and Exam F1 (+44.06, 95\% CI [41.26, 46.77], $p<0.0001$). PACT also improves over its own Qwen3-8B base model by +22.78 Strict DA points (Table~\ref{tab:cross_arch}). Notably, GPT-4.1 attains the highest symptom coverage (94.04\%), slightly exceeding PACT (93.39\%); this difference is not statistically significant ($p=0.061$), indicating that symptom gathering ability alone does not substitute for the structured diagnostic reasoning that PACT acquires through multi-paradigm consensus training.

Many medical-specialized LLMs are optimized for single-turn QA or static reasoning rather than interactive diagnosis \cite{lai2026doctorr1,feng2025doctoragentrl}. Doctor-R1, which benefits from multi-turn interaction training, becomes the strongest medical-specialized baseline. Mixed-SFT on DPS data already surpasses Doctor-R1 on Qwen3-8B (51.48\% vs.\ 32.65\%; Table~\ref{tab:cross_arch}), and PACT further improves over Mixed-SFT by +3.82 Strict DA points with statistical significance (95\% CI [1.31, 6.28], $p=0.003$), while also achieving higher Exam F1 (61.33 vs.\ 59.83).

\noindent\textbf{English-translated setting.} To test whether the learned diagnostic behavior transfers beyond the original Chinese setting, we also evaluate an English-translated variant. English PACT improves over the English Base by +12.45 Strict DA, +18.65 Exam F1, and reduces average dialogue rounds from 3.73 to 1.27 (Appendix Table~\ref{tab:english_translation}).

\subsection{Ablation Study}

\begin{table}[!t]
    \centering
    \resizebox{\columnwidth}{!}{
    \begin{tabular}{llccc}
        \toprule
        Base Model & Config & Strict DA & Exam F1 & Symp Cov \\
        \midrule
        \multirow{3}{*}{Qwen3.5-9B}
            & Base      & 35.90 & 15.00 & 78.15 \\
            & Mixed-SFT & 48.29 & 62.45 & \textbf{93.52} \\
            & PACT      & \textbf{49.76} & \textbf{63.11} & 79.52 \\
        \midrule
        \multirow{3}{*}{Qwen3-8B}
            & Base      & 32.53 & 21.53 & 77.11 \\
            & Mixed-SFT & 51.48 & 59.83 & \textbf{93.72} \\
            & PACT      & \textbf{55.31} & \textbf{61.33} & 93.39 \\
        \midrule
        \multirow{3}{*}{Llama3.1-8B}
            & Base      & 16.89 & 16.68 & 75.04 \\
            & Mixed-SFT & 51.37 & 45.87 & 74.79 \\
            & PACT      & \textbf{53.03} & \textbf{60.92} & \textbf{93.14} \\
        \midrule
        \multirow{3}{*}{Ministral-3-8B}
            & Base      & 20.94 & 5.08 & 43.28 \\
            & Mixed-SFT & 36.44 & 28.61 & \textbf{88.54} \\
            & PACT      & \textbf{50.26} & \textbf{45.71} & 75.38 \\
        \bottomrule
    \end{tabular}
    }
    \caption{Cross-architecture generalization (key metrics). Full results in Appendix Table~\ref{tab:cross_arch_full}.}
    \label{tab:cross_arch}
    \vspace{-2mm}
\end{table}

\begin{table}[htbp]
    \centering
    \resizebox{\columnwidth}{!}{
    \begin{tabular}{lccc}
        \toprule
        Setting & Strict DA & Exam F1 & Symp Cov \\
        \midrule
        All four paradigms & 51.48 & \textbf{59.83} & \textbf{93.39} \\
        \midrule
        w/o Serial    & 51.43 & 54.51 & 93.21 \\
        w/o Tree      & 50.34 & 53.87 & 93.22 \\
        w/o Heap      & 50.23 & 53.64 & 92.69 \\
        w/o Feedback  & \textbf{51.91} & 53.29 & 92.94 \\
        \bottomrule
    \end{tabular}
    }
    \caption{Leave-one-out data-paradigm ablation under SFT on Qwen3-8B. Each row removes one reasoning-paradigm subset from the training data.}
    \vspace{-2mm}
    \label{tab:abl_paradigm}
\end{table}

\noindent\textbf{Leave-one-out data-paradigm ablation.} Table~\ref{tab:abl_paradigm} shows that SFT with all four reasoning-paradigm subsets achieves the strongest exam recommendation quality and the highest symptom coverage. Removing any paradigm reduces Exam F1 by 5.3--6.5 points, while the leave-one-out variants remain competitive on Strict DA and Symp Cov. This suggests that the four reasoning paradigms provide complementary supervision rather than any single paradigm being indispensable.

\begin{table}[htbp]
    \centering
    \resizebox{\columnwidth}{!}{
    \begin{tabular}{lccc}
        \toprule
        Strategy & Strict DA & Exam F1 & Symp Cov \\
        \midrule
        Mixed-SFT & 51.48 & 59.83 & 93.72 \\
        Sequential Curriculum & 50.17 & 57.86 & \textbf{94.72} \\
        FedAvg-style Avg & 54.39 & 18.74 & 94.11 \\
        Indep. + Linear & 52.91 & 56.27 & 93.92 \\
        Indep. + TIES & 40.41 & 59.03 & 89.81 \\
        Indep. + DARE-TIES & 52.45 & 57.16 & 93.66 \\
        \midrule
        PACT & \textbf{55.31}$^*$ & \textbf{61.33} & 93.39 \\
        \bottomrule
    \end{tabular}
    }
    \caption{Training paradigm ablation on Qwen3-8B. The table compares mixed training, sequential learning, post-hoc merging \cite{ilharco2023task,yadav2023ties}, periodic averaging, and PACT. $^*$ indicates a significant improvement over Mixed-SFT under paired bootstrap over EMR cases ($p<0.05$).}
    \vspace{-2mm}
    \label{tab:abl_strategy}
\end{table}

\noindent\textbf{Training paradigm ablation.} Table~\ref{tab:abl_strategy} shows that PACT achieves the best Strict DA and significantly improves over Mixed-SFT (+3.82, $p=0.003$). PACT also outperforms Mixed-SFT on Exam F1 and surpasses Sequential Curriculum on both Strict DA (+5.14) and Exam F1 (+3.47), while Sequential Curriculum retains the highest symptom coverage. Post-hoc branch merging is sensitive to the merge operator: TIES degrades Strict DA to 40.41\%, and DARE-TIES only partially recovers performance; detailed post-hoc merging results are reported in Appendix Table~\ref{tab:posthoc_details}.

\begin{table}[htbp]
    \centering
    \resizebox{\columnwidth}{!}{
    \begin{tabular}{lccc}
        \toprule
        Regularization & Strict DA & Exam F1 & Symp Cov \\
        \midrule
        None & 33.45 & 56.96 & 89.83 \\
        L2 Prox & 33.16 & 58.00 & 89.23 \\
        Cosine & 54.05 & \textbf{62.32} & 91.02 \\
        L1 Anchor & \textbf{55.31} & 61.33 & \textbf{93.39} \\
        \bottomrule
    \end{tabular}
    }
    \caption{Regularization distance ablation with TIES aggregation fixed. L2 Prox follows the FedProx proximal objective \cite{li2020fedprox}.}
    \label{tab:abl_regularization}
    \vspace{-2mm}
\end{table}

\noindent\textbf{Regularization distance ablation.} As shown in Table~\ref{tab:abl_regularization}, L1 Anchor regularization gives the strongest Strict DA and symptom coverage, whereas no regularization or L2 proximity leads to unstable anchor training. This supports L1 shrinkage to keep Branches close to the Anchor while allowing sparse paradigm-specific deviations.

\begin{table}[htbp]
    \centering
    \resizebox{\columnwidth}{!}{
    \begin{tabular}{lccc}
        \toprule
        Aggregation & Strict DA & Exam F1 & Symp Cov \\
        \midrule
        Linear Avg & 55.19 & 33.74 & 93.28 \\
        SLERP & 54.45 & 36.43 & 93.68 \\
        TIES & \textbf{55.31} & \textbf{61.33} & \textbf{93.39} \\
        \bottomrule
    \end{tabular}
    }
    \caption{Aggregation operator ablation with L1 Anchor regularization fixed. Methods: Linear Avg, SLERP \cite{goddard2024arcee}, and TIES \cite{yadav2023ties}.}
    \label{tab:abl_aggregation}
    \vspace{-3mm}
\end{table}

\noindent\textbf{Aggregation operator ablation.} Table~\ref{tab:abl_aggregation} shows that with L1 Anchor regularization fixed, TIES maintains both diagnostic accuracy and examination quality, while Linear Avg and SLERP preserve Strict DA but reduce Exam F1. This indicates that sign-consensus filtering is important for resolving conflicting paradigm updates.

\noindent\textbf{Cross-architecture generalization.} Table~\ref{tab:cross_arch} shows that PACT improves over the zero-shot Base across all architectures, and consistently outperforms Mixed-SFT on Strict DA: Qwen3-8B (+3.83), Qwen3.5-9B (+1.47), Llama3.1-8B (+1.66), and Ministral-3-8B (+13.82). PACT also substantially improves examination planning quality on Llama3.1-8B (+15.05 Exam F1) and Ministral-3-8B (+17.10 Exam F1). Full results are in Appendix Table~\ref{tab:cross_arch_full}; DPS data-quality analysis is reported in Appendix~\ref{sec:additional_results}.

\section{Conclusion}

We present PACT, a framework for multi-turn medical diagnostic dialogue that combines privileged data synthesis with training-time Branch aggregation. DPS constructs validated dialogues under multiple reasoning paradigms while preserving doctor-patient information asymmetry, and PACT aggregates paradigm-specific LoRA Branches into a single deployable Anchor through periodic sign-consensus aggregation. Experiments on a dynamic Chinese medical diagnosis benchmark show that PACT achieves state-of-the-art performance among compared task-adapted, post-hoc merging, and zero-shot LLM baselines under the same simulator-based evaluation protocol.

\section*{Limitations}

This study remains limited by its reliance on synthetic supervision, LLM-based patient simulation and judging, and de-identified EMRs from a single Internal Medicine department. Recurring failure modes---including premature closure, examination omission, and over-broad differentials---are discussed in Appendix~\ref{sec:failure_modes}. The results should therefore be interpreted as evidence for task-specific diagnostic-dialogue adaptation rather than clinical deployment readiness. Public models and baselines are used under their original licenses or service terms, and proprietary LLMs are accessed through their official services. The de-identified internal EMRs and derived DPS/benchmark artifacts are restricted to research use and are not intended for clinical deployment or unrestricted redistribution. Future work should incorporate clinician evaluation, judge-robustness checks, broader cross-institutional validation, and safer evaluation protocols.

\bibliography{custom}

\appendix

\section{Prompt Templates and Generation Controls}
\label{sec:prompt_templates}

All DPS synthesis prompts are written in Chinese to match the source EMRs and the target diagnostic dialogue setting. For readability, we provide faithful English renderings of the complete prompt structures below while preserving the original placeholders.

\paragraph{Shared Doctor instruction.}
You are an experienced and logically rigorous doctor conducting an online consultation. Your goal is to collect information through multi-turn interaction and, before the final turn, provide both examination recommendations and a preliminary diagnosis. Examination recommendations include laboratory tests, imaging tests, and other auxiliary examinations. The preliminary diagnosis should be based on the chief complaint, history, physical signs, and available examinations; because information is incomplete, it should be framed as a preliminary rather than final diagnosis. Information should be organized in a medical-record format covering basic information, chief complaint, history of present illness, past medical history, allergy history, and physical examination. The Doctor receives the dialogue history \texttt{\{\{history\}\}} and the latest patient message \texttt{\{\{message\}\}}, reasons internally, and outputs only the direct patient-facing reply.

\paragraph{Serial Doctor template.}
The Doctor follows a step-by-step consultation policy. The internal reasoning template contains: (1) known information organized by medical-record fields, (2) diagnostic hypotheses matching the current evidence, (3) missing information to be collected, (4) a decision to either collect more information or give the final opinion, and (5) a response strategy. This template encourages linear hypothesis testing and asks one or two high-priority questions per turn.

\paragraph{Tree Doctor template.}
The Doctor maintains a set of candidate diseases and uses high-discrimination questions to narrow the differential diagnosis. The internal reasoning template contains: (1) known information, (2) current suspected diseases, (3) discriminative features among these diseases, (4) the best question for maximizing information gain, (5) a collect-or-finalize decision, and (6) a response strategy.

\paragraph{Heap Doctor template.}
The Doctor prioritizes information by diagnostic importance, urgency, and specificity. The internal reasoning template contains: (1) known information, (2) analysis of known symptoms with high/medium/low importance weights, (3) the most important symptoms to lock onto, (4) deeper inquiry over missing high-value evidence, (5) a collect-or-finalize decision, and (6) a response strategy.

\paragraph{Feedback Doctor template.}
The Doctor performs a generate--critique--revise rehearsal before responding. The internal reasoning template contains: (1) known information, (2) an initial response plan, (3) error detection for logical gaps, insufficient evidence, or previously denied symptoms, (4) control revision, and (5) the final decision. This template encourages reflection before producing the patient-facing reply.

\paragraph{Patient template.}
The Patient is instructed to role-play an online consultation patient using only patient-visible facts \texttt{\{\{patient\_memory\}\}} and the dialogue history \texttt{\{\{history\}\}}. If the history is empty, the Patient starts with a short natural opening based only on the chief complaint. The Patient must convert medical-record language into lay expressions, answer only what the Doctor asks, avoid disclosing the full record at once, and answer ``not sure'', ``did not notice'', or ``have not done that test'' when the requested information is absent from the visible memory.

\paragraph{Doctor-Supervisor template.}
The Doctor-Supervisor acts as a privileged clinical supervisor with access to the full ground-truth EMR \texttt{\{\{ground\_truth\}\}}, target diagnosis \texttt{\{\{diagnosis\}\}}, dialogue history \texttt{\{\{history\}\}}, and Doctor draft \texttt{\{\{draft\_response\}\}}. It checks whether the Doctor should receive PASS or REWRITE. Required rewrites include severe diagnostic drift, missed high-risk clues, logical dead ends, missing key examinations at closure, failure to use already collected evidence, over-suspicion, and over-examination. The Supervisor must preserve knowledge isolation: it may use privileged information to judge the process, but its rewritten patient-facing response must not reveal facts that the Doctor could not know from the dialogue. Rewrites are limited to adding necessary questions, removing low-value questions, or appending missing examination suggestions.

\paragraph{Patient-Supervisor template.}
The Patient-Supervisor receives patient-visible information \texttt{\{\{patient\_profile\}\}}, dialogue history \texttt{\{\{history\}\}}, and the Patient draft \texttt{\{\{draft\_response\}\}}. It checks whether the Patient response faithfully follows the visible facts. Required rewrites include fabricated positive symptoms, direct copying of medical jargon, over-answering beyond the Doctor's question, and role-play inconsistencies. The output contains a correction analysis, a PASS-or-REWRITE decision, and a concise rewritten patient reply only when rewriting is necessary.

\section{Dataset, Evaluation, and Hyperparameter Details}
\label{sec:data_eval_details}

\subsection{Dataset Statistics}

Table~\ref{tab:dataset_stats} provides the dataset statistics used in our experiments. All splits are performed at the EMR level to avoid overlap between training and evaluation cases.

\begin{table}[!t]
    \centering
    \resizebox{\columnwidth}{!}{
    \begin{tabular}{lc}
        \toprule
        Item & Value \\
        \midrule
        De-identified EMR pool & 10,000 \\
        Clinical department & Internal Medicine \\
        EMRs sampled for DPS synthesis & 6,357 \\
        Raw synthesized dialogues & 25,428 \\
        Retained unique EMRs after filtering & 5,171 \\
        Retained synthesized dialogues & 20,684 \\
        Reasoning paradigms per EMR & 4 \\
        Train / validation / test split & 85\% / 5\% / 10\% \\
        Unique diagnostic labels in retained pool & 145 \\
        Held-out dynamic evaluation EMRs & 438 \\
        Rollouts per held-out EMR & 4 \\
        Total dynamic evaluation episodes & 1,752 \\
        Maximum dialogue rounds & 8 \\
        \bottomrule
    \end{tabular}
    }
    \caption{Dataset statistics for DPS synthesis and dynamic evaluation.}
    \label{tab:dataset_stats}
\end{table}

\subsection{Evaluation Rubric}

Table~\ref{tab:evaluation_rubric} summarizes the metrics used by the GPT-4o judge in dynamic multi-turn evaluation. The judge evaluates both outcome quality and consultation process quality after each completed dialogue.

\begin{table*}[!t]
    \centering
    \resizebox{\textwidth}{!}{
    \begin{tabular}{lll}
        \toprule
        Dimension & Metric & Definition \\
        \midrule
        Diagnosis & Strict DA & The top-1 diagnosis exactly matches the ground-truth diagnosis or an accepted synonym. \\
        Diagnosis & Relaxed DA & Credits strict matches, hierarchical matches, or top-3 diagnostic hits. \\
        Diagnosis & Soft Score & Assigns 1.0 / 0.7 / 0.3 / 0 for strict, hierarchical, top-3, and missed diagnoses. \\
        Examination & Exam Precision & Fraction of recommended examinations that match the gold-standard examination list. \\
        Examination & Exam Recall & Fraction of gold-standard examinations covered by the model's recommendations. \\
        Examination & Exam F1 & Harmonic mean of Exam Precision and Exam Recall. \\
        Process & Symptom Coverage & Coverage of key clinical information collected during the dialogue. \\
        Efficiency & Avg Rnd & Average number of dialogue rounds before termination. \\
        \bottomrule
    \end{tabular}
    }
    \caption{Evaluation rubric for dynamic multi-turn diagnosis.}
    \label{tab:evaluation_rubric}
\end{table*}

\subsection{Training and Evaluation Hyperparameters}

Table~\ref{tab:training_hparams} lists the main training and evaluation hyperparameters. Unless otherwise specified, the same settings are used for the primary Qwen3-8B experiments and cross-architecture runs.

\begin{table}[!t]
    \centering
    \resizebox{\columnwidth}{!}{
    \begin{tabular}{lc}
        \toprule
        Hyperparameter & Value \\
        \midrule
        LoRA rank $r$ & 64 \\
        LoRA $\alpha$ & 64 \\
        LoRA target modules & Attention projection matrices \\
        Optimizer & AdamW \\
        Learning rate & $5 \times 10^{-4}$ \\
        Schedule & Cosine \\
        Warmup ratio & 0.03 \\
        Precision & bfloat16 \\
        Effective batch size & 64 \\
        Number of GPUs & 4 \\
        PACT synchronization strength $\lambda$ & 200 \\
        Aggregation interval $T^*$ & 0.25 epochs \\
        TIES density $\rho$ & 0.5 \\
        Maximum training sequence length & 4096 \\
        Dynamic evaluation episodes per EMR & 4 \\
        Dynamic evaluation maximum rounds & 8 \\
        \bottomrule
    \end{tabular}
    }
    \caption{Main hyperparameters used for training and evaluation.}
    \label{tab:training_hparams}
\end{table}

\paragraph{Preprocessing.}
Dialogue data are converted into chat-style training examples by mapping Patient utterances to user turns and Doctor utterances to assistant turns. Loss is applied only to assistant responses. Residual thinking tags are removed before training, and models without a built-in chat template are assigned a ChatML-style template for consistent formatting.

\subsection{Dataset Filtering Analysis}

Table~\ref{tab:filtering_analysis} summarizes the filtering stages used to construct the final DPS training pool. We first remove very short or invalid dialogue trajectories, and then apply record-level quality filtering: if at least two paradigm-specific dialogues from the same EMR fail diagnosis matching, all four dialogues from that EMR are removed. This rule preserves paradigm balance in the retained set, but may bias the training pool toward cases with clearer diagnostic trajectories.

\begin{table}[!t]
    \centering
    \resizebox{\columnwidth}{!}{
    \begin{tabular}{lcc}
        \toprule
        Stage & EMRs & Dialogues \\
        \midrule
        Raw DPS synthesis & 6,357 & 25,428 \\
        Round/validity filtered & 6,128 & 24,512 \\
        Record-level quality filtered & 5,171 & 20,684 \\
        \bottomrule
    \end{tabular}
    }
    \caption{DPS filtering stages for the synthesis-derived training pool.}
    \label{tab:filtering_analysis}
\end{table}

The record-level quality filter removes 957 EMRs from the round-filtered pool, corresponding to 3,828 paradigm-specific dialogues. Among removed EMRs, 685 have two failed paradigms, 217 have three failed paradigms, and 55 fail across all four paradigms. We therefore treat filtering as a quality-control step rather than a neutral sampling operation, and explicitly acknowledge its potential easy-case bias in the paper limitations.

\section{Theoretical Discussion}
\label{sec:theory_appendix}

This section gives a more detailed justification of the specialization--consensus design in PACT. The goal is not to establish a global convergence theorem for non-convex LLM training, but to clarify why periodic consensus can be preferable to either mixed SFT or post-hoc branch merging under standard local assumptions.

\subsection{Local Assumptions}

Let $\Delta^{(k)}$ denote the LoRA update learned by Branch $k$ on paradigm-specific data $D_k$, and let $\mathcal{L}_k(\Delta)$ be its local training objective in the LoRA update space. We use two local assumptions.

\paragraph{Local PL condition.}
Around the optimization trajectory of each Branch, assume $\mathcal{L}_k$ satisfies a Polyak--\L ojasiewicz (PL) inequality with parameter $\mu_k > 0$:
\begin{equation}
\label{eq:appendix_pl}
\|\nabla \mathcal{L}_k(\Delta)\|_2^2 \geq 2\mu_k \left(\mathcal{L}_k(\Delta)-\mathcal{L}_k(\Delta_k^*)\right),
\end{equation}
where $\Delta_k^*$ is a local minimizer for Branch $k$. This assumption is weaker than local strong convexity and is used only to reason about local descent and bounded suboptimality.

\paragraph{Local merge compatibility.}
Because all Branches start from the same base model and optimize low-rank LoRA updates, we assume that a local low-loss region connects compatible Branch updates. Formally, for sign-consistent updates $\Delta^{(i)}$ and $\Delta^{(j)}$, there exists a path $\gamma(\alpha)$ between them such that $\mathcal{L}_k(\gamma(\alpha))$ does not exceed the endpoint losses by more than a small compatibility error $\epsilon_{\mathrm{lmc}}$. This is a restricted-subspace version of local linear mode connectivity and motivates aggregating LoRA deltas rather than full model weights.

\subsection{Why Simple Averaging Can Lose Signal}

Consider a single parameter coordinate $j$ across $K$ Branch updates. Simple averaging produces
\begin{equation}
\label{eq:appendix_avg}
\bar{\Delta}_j = \frac{1}{K}\sum_{k=1}^{K}\Delta^{(k)}_j.
\end{equation}
If Branches trained on different reasoning paradigms update this coordinate in opposite directions, a coordinate-level analogue of the gradient conflicts studied in multi-task learning \cite{yu2020pcgrad,liu2021cagrad}, positive and negative values cancel. Let $P_j=\{k:\Delta^{(k)}_{j}>0\}$ and $N_j=\{k:\Delta^{(k)}_{j}<0\}$. When both sets are non-empty, the magnitude $|\bar{\Delta}_j|$ can be much smaller than the average magnitude of either sign-consistent group. Thus, averaging may remove a useful paradigm-shared signal simply because a minority of Branches moves in the opposite direction.

TIES aggregation reduces this cancellation by retaining the majority sign:
\begin{equation}
\label{eq:appendix_ties}
\Delta^{\mathrm{ties}}_j =
\begin{cases}
\frac{1}{|P_j|}\sum_{k\in P_j}\Delta^{(k)}_j, & |P_j|>|N_j|, \\
\frac{1}{|N_j|}\sum_{k\in N_j}\Delta^{(k)}_j, & |N_j|\geq |P_j|.
\end{cases}
\end{equation}
Low-magnitude coordinates are trimmed before sign election, so the aggregation focuses on dimensions with stronger evidence. This does not guarantee that the majority direction is always correct, but it prevents direct destructive cancellation and provides an explicit filter for conflicting updates.

\subsection{Effect of L1 Anchor Regularization}

PACT trains each Branch with the local objective
\begin{equation}
\label{eq:appendix_l1_objective}
\widetilde{\mathcal{L}}_k(\Delta)=\mathcal{L}_k(\Delta)+\lambda\|\Delta-\Delta^{\mathrm{anc}}\|_1.
\end{equation}
At a stationary point $\widehat{\Delta}_k$ of the regularized objective, the first-order optimality condition gives
\begin{equation}
\label{eq:appendix_stationary}
0 \in \nabla\mathcal{L}_k(\widehat{\Delta}_k)+\lambda\partial\|\widehat{\Delta}_k-\Delta^{\mathrm{anc}}\|_1.
\end{equation}
Therefore, there exists a subgradient $g_k\in\partial\|\widehat{\Delta}_k-\Delta^{\mathrm{anc}}\|_1$ such that $\nabla\mathcal{L}_k(\widehat{\Delta}_k)=-\lambda g_k$. Since each coordinate of an L1 subgradient has magnitude at most one, $\|g_k\|_2^2\leq p$, where $p$ is the number of trainable LoRA parameters. Combining this with the PL condition yields
\begin{equation}
\label{eq:appendix_bound}
\mathcal{L}_k(\widehat{\Delta}_k)-\mathcal{L}_k(\Delta_k^*)
\leq \frac{\lambda^2\|g_k\|_2^2}{2\mu_k}
\leq \frac{\lambda^2 p}{2\mu_k}.
\end{equation}
This is a conservative upper bound. In practice, the effective support of $g_k$ is often much smaller than $p$ because many coordinates either remain close to the Anchor or are trimmed during aggregation. The bound therefore supports a qualitative conclusion rather than a tight prediction: moderate $\lambda$ controls Branch drift with a bounded local cost, while overly large $\lambda$ can suppress specialization.

\subsection{Why Periodic Consensus Helps}

The aggregation interval $T^*$ determines how long Branches can specialize before returning to consensus. If $T^*$ is too small, Branches are repeatedly pulled back before learning useful paradigm-specific behavior. If $T^*$ is too large, Branch updates may become less merge-compatible, analogous to client drift under heterogeneous federated optimization \cite{karimireddy2020scaffold}, making post-hoc aggregation unstable. PACT alternates between local exploration and Anchor aggregation, aiming to keep Branches within a merge-compatible neighborhood while still allowing each paradigm to contribute specialized information. This explains the empirical sensitivity to $\lambda$ and $T^*$ observed in Section~\ref{sec:experiments}.

\section{PACT Implementation Details}
\label{sec:pact_impl}

\paragraph{Materialized LoRA deltas.}
For each LoRA module, we compute the effective update as the materialized product $\Delta = BA$. The Anchor regularization in Eq.~\ref{eq:total_loss} is applied to these materialized deltas, and gradients are back-propagated to the low-rank factors $B$ and $A$.

\paragraph{Rank projection.}
Sign-consensus aggregation is performed on materialized Branch deltas. Because the coordinate-wise aggregated matrix is generally full-rank, the result is projected back to the original LoRA rank $r$ through truncated SVD before it is stored as the updated Anchor.

\paragraph{Memory-efficient regularization.}
Computing the L1 Anchor penalty over all LoRA modules in one graph can increase peak memory. We therefore compute and back-propagate the regularization term module by module, freeing each intermediate graph immediately after its backward pass. This preserves the same objective while reducing peak memory in implementation.

\paragraph{Computational overhead.}
PACT introduces training-time overhead from maintaining four Branch adapters, computing the L1 Anchor penalty, and running periodic sign-consensus aggregation. The aggregation is performed only every $T^*$ epochs and operates on low-rank LoRA deltas rather than full model weights, so it is not the dominant cost compared with forward and backward passes. At inference time, only the Anchor is deployed, making PACT identical to a standard single-LoRA model in parameter count and latency.

\section{Statistical Significance Tests}
\label{sec:significance_tests}

We conduct paired bootstrap significance tests over EMR cases rather than over individual dialogue episodes. For each model and metric, we first average the four independent episodes belonging to the same held-out EMR, yielding 438 paired case-level scores. We then draw 20,000 bootstrap samples over EMR cases and compute two-sided $p$-values from the bootstrap distribution of paired differences. Table~\ref{tab:significance} reports the main comparisons; Delta denotes PACT minus the comparison method.

\begin{table}[!t]
    \centering
    \resizebox{\columnwidth}{!}{
    \begin{tabular}{llccc}
        \toprule
        Comparison & Metric & Delta & 95\% CI & $p$-value \\
        \midrule
        PACT--Mixed-SFT & Strict DA & +3.82 & [1.31, 6.28] & 0.003 \\
        PACT--Mixed-SFT & Soft Score & +1.28 & [-0.65, 3.21] & 0.198 \\
        PACT--Mixed-SFT & Exam F1 & +1.50 & [-0.25, 3.26] & 0.093 \\
        PACT--Sequential & Strict DA & +5.14 & [2.71, 7.56] & $<0.001$ \\
        PACT--Sequential & Exam F1 & +3.47 & [1.68, 5.27] & 0.001 \\
        PACT--GPT-4.1 & Strict DA & +18.61 & [15.24, 22.03] & $<0.0001$ \\
        PACT--GPT-4.1 & Soft Score & +11.30 & [8.68, 13.94] & $<0.0001$ \\
        PACT--GPT-4.1 & Exam F1 & +44.06 & [41.26, 46.77] & $<0.0001$ \\
        PACT--GPT-4.1 & Symp Cov & -0.64 & [-1.31, 0.03] & 0.061 \\
        \bottomrule
    \end{tabular}
    }
    \caption{Paired bootstrap significance tests over EMR cases. Delta denotes PACT minus the comparison method.}
    \label{tab:significance}
\end{table}

\section{Post-hoc Merge Details}
\label{sec:posthoc_details}

Table~\ref{tab:posthoc_details} reports the detailed post-hoc merging results used to summarize the independent-branch baseline in Table~\ref{tab:abl_strategy}. Different merge operators favor different metrics, indicating that post-hoc merging is sensitive to the aggregation rule.

\begin{table}[!t]
    \centering
    \resizebox{\columnwidth}{!}{
    \begin{tabular}{lccc}
        \toprule
        Method & Strict DA & Exam F1 & Symp Cov \\
        \midrule
        Indep. + Linear & 52.91 & 56.27 & 93.92 \\
        Indep. + TIES & 40.41 & 59.03 & 89.81 \\
        Indep. + DARE-TIES & 52.45 & 57.16 & 93.66 \\
        \bottomrule
    \end{tabular}
    }
    \caption{Post-hoc merge details for independently trained Qwen3-8B Branches.}
    \label{tab:posthoc_details}
\end{table}

\section{Qualitative Failure Modes}
\label{sec:failure_modes}

Although PACT improves diagnosis, examination recommendation, and symptom coverage, dynamic evaluation still reveals several recurring failure modes. \textbf{Premature closure} occurs when the agent provides a diagnosis before collecting enough discriminative evidence. \textbf{Examination omission} occurs when the final response contains a plausible diagnosis but misses key auxiliary examinations. \textbf{Over-broad differential diagnosis} occurs when the agent lists many possible diseases without committing to the most supported diagnosis. \textbf{Simulator and judge dependence} remains a limitation because both the patient simulator and outcome judge are LLM-based. These failure categories motivate future work on clinician review and safer clinical evaluation protocols.

\section{Analysis and Additional Experimental Results}
\label{sec:additional_results}

\begin{table*}[!t]
    \centering
    \resizebox{\textwidth}{!}{
    \begin{tabular}{llc ccc ccc cc}
        \toprule
        \multirow{2}{*}{Base Model} & \multirow{2}{*}{Config} & \multirow{2}{*}{Size} & \multicolumn{3}{c}{Diagnosis} & \multicolumn{3}{c}{Examination} & \multicolumn{2}{c}{Process} \\
        \cmidrule(lr){4-6} \cmidrule(lr){7-9} \cmidrule(lr){10-11}
        & & & Strict DA & Relaxed DA & Soft Score & Exam F1 & Exam P & Exam R & Avg Rnd & Symp Cov \\
        \midrule
        \multirow{3}{*}{Qwen3.5-9B}
            & Base      & 9B & 35.90 & 71.06 & 59.24 & 15.00 & 16.79 & 16.70 & 5.78 & 78.15 \\
            & Mixed-SFT & 9B & 48.29 & 83.33 & 72.96 & 62.45 & 69.62 & 61.97 & 4.24 & 93.52 \\
            & PACT      & 9B & 49.76 & 82.23 & 71.73 & 63.11 & 64.58 & 68.53 & 2.06 & 79.52 \\
        \midrule
        \multirow{3}{*}{Qwen3-8B}
            & Base      & 8B & 32.53 & 73.97 & 59.03 & 21.53 & 23.86 & 23.95 & 4.70 & 77.11 \\
            & Mixed-SFT & 8B & 51.48 & 86.02 & 74.88 & 59.83 & 60.42 & 63.17 & 4.17 & 93.72 \\
            & PACT      & 8B & 55.31 & 85.79 & 76.16 & 61.33 & 57.50 & 74.27 & 4.02 & 93.39 \\
        \midrule
        \multirow{3}{*}{Llama3.1-8B}
            & Base      & 8B & 16.89 & 70.03 & 50.94 & 16.68 & 18.40 & 17.46 & 3.77 & 75.04 \\
            & Mixed-SFT & 8B & 51.37 & 86.74 & 77.81 & 45.87 & 49.64 & 47.48 & 2.56 & 74.79 \\
            & PACT      & 8B & 53.03 & 84.19 & 74.54 & 60.92 & 66.82 & 61.15 & 4.27 & 93.14 \\
        \midrule
        \multirow{3}{*}{Ministral-3-8B}
            & Base      & 8B & 20.94 & 52.34 & 41.83 & 5.08 & 6.43 & 5.16 & 2.71 & 43.28 \\
            & Mixed-SFT & 8B & 36.44 & 78.53 & 64.35 & 28.61 & 32.90 & 30.93 & 3.69 & 88.54 \\
            & PACT      & 8B & 50.26 & 78.39 & 69.51 & 45.71 & 49.03 & 48.33 & 2.00 & 75.38 \\
        \bottomrule
    \end{tabular}
    }
    \caption{Full cross-architecture generalization results. Each model shows three configurations: zero-shot Base, Mixed-SFT (standard training on DPS data), and PACT.}
    \label{tab:cross_arch_full}
\end{table*}

\subsection{DPS Data Quality Analysis}

\begin{table}[!t]
    \centering
    \resizebox{\columnwidth}{!}{
    \begin{tabular}{lccccc}
        \toprule
        Setting & Leak. $\downarrow$ & Invalid $\downarrow$ & Diag. $\uparrow$ & Exam $\uparrow$ & Rewrite \\
        \midrule
        Single-model & 18.7 & 24.5 & 71.2 & 62.8 & -- \\
        Doctor-Patient & 6.8 & 17.9 & 78.4 & 68.5 & -- \\
        w/o Patient-S & 4.9 & 11.8 & 84.6 & 74.3 & 13.2 \\
        w/o Doctor-S & 7.6 & 9.4 & 81.5 & 70.9 & 9.7 \\
        DPS full & \textbf{2.1} & \textbf{5.8} & \textbf{91.3} & \textbf{82.6} & 16.4 \\
        \bottomrule
    \end{tabular}
    }
    \caption{DPS data quality and Supervisor ablation. Leakage and invalid rates are lower-is-better; diagnosis and examination match are higher-is-better.}
    \label{tab:dps_quality}
\end{table}

DPS substantially reduces information leakage and invalid dialogue behaviors while improving diagnosis and examination alignment. The Supervisor Rewrite Rate remains moderate under the full setting, supporting the minimal-intervention design: privileged supervision corrects clear process failures without fully overriding the natural doctor-patient simulation.

\subsection{Hyperparameter Sensitivity}

\begin{figure}[t]
    \centering
    \includegraphics[width=\columnwidth]{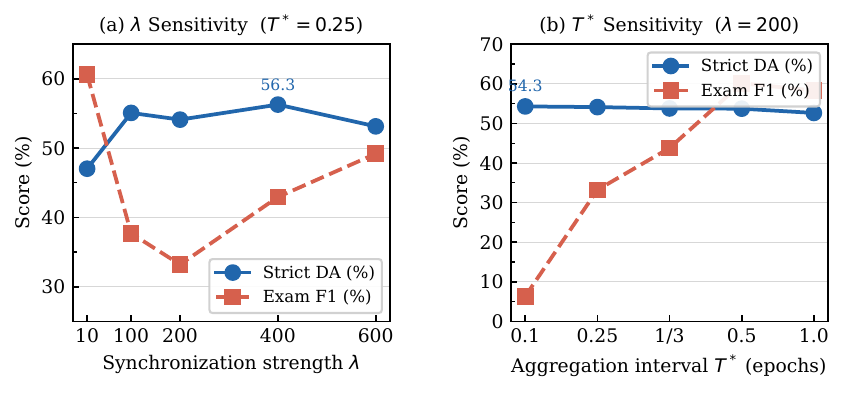}
    \caption{Hyperparameter sensitivity analysis. (a) Effect of synchronization strength $\lambda$. (b) Effect of aggregation interval $T^*$.}
    \label{fig:hyperparam}
\end{figure}

We vary the synchronization strength $\lambda \in \{10, 100, 200, 400, 600\}$ and aggregation interval $T^* \in \{0.1, 0.25, 0.5, 1.0\}$ epochs. Small $\lambda$ values make PACT approach independent training, while overly large values suppress specialization; performance peaks around $\lambda = 200$. Similarly, overly frequent aggregation constrains Branch exploration too early, whereas infrequent aggregation allows excessive drift before merging. The default $T^* = 0.25$ epochs provides the best overall balance.

\begin{table}[!t]
    \centering
    \resizebox{\columnwidth}{!}{
    \begin{tabular}{lccc}
        \toprule
        TIES density $\rho$ & Strict DA & Exam F1 & Symp Cov \\
        \midrule
        0.25 & 53.94 & 59.72 & 92.88 \\
        0.50 & \textbf{55.31} & \textbf{61.33} & \textbf{93.39} \\
        0.75 & 54.62 & 60.41 & 93.18 \\
        1.00 & 53.78 & 58.96 & 92.74 \\
        \bottomrule
    \end{tabular}
    }
    \caption{Sensitivity to TIES density $\rho$ on Qwen3-8B, with $\lambda{=}200$ and $T^*{=}0.25$ fixed.}
    \label{tab:rho_sensitivity}
\end{table}

Table~\ref{tab:rho_sensitivity} shows that moderate trimming works best. A small density removes useful Branch-specific signal, while no trimming retains low-magnitude noisy coordinates and weakens sign-consensus aggregation. We therefore use $\rho=0.5$ as the default.

\subsection{Parameter Drift Analysis}

\begin{figure}[t]
    \centering
    \includegraphics[width=0.98\columnwidth]{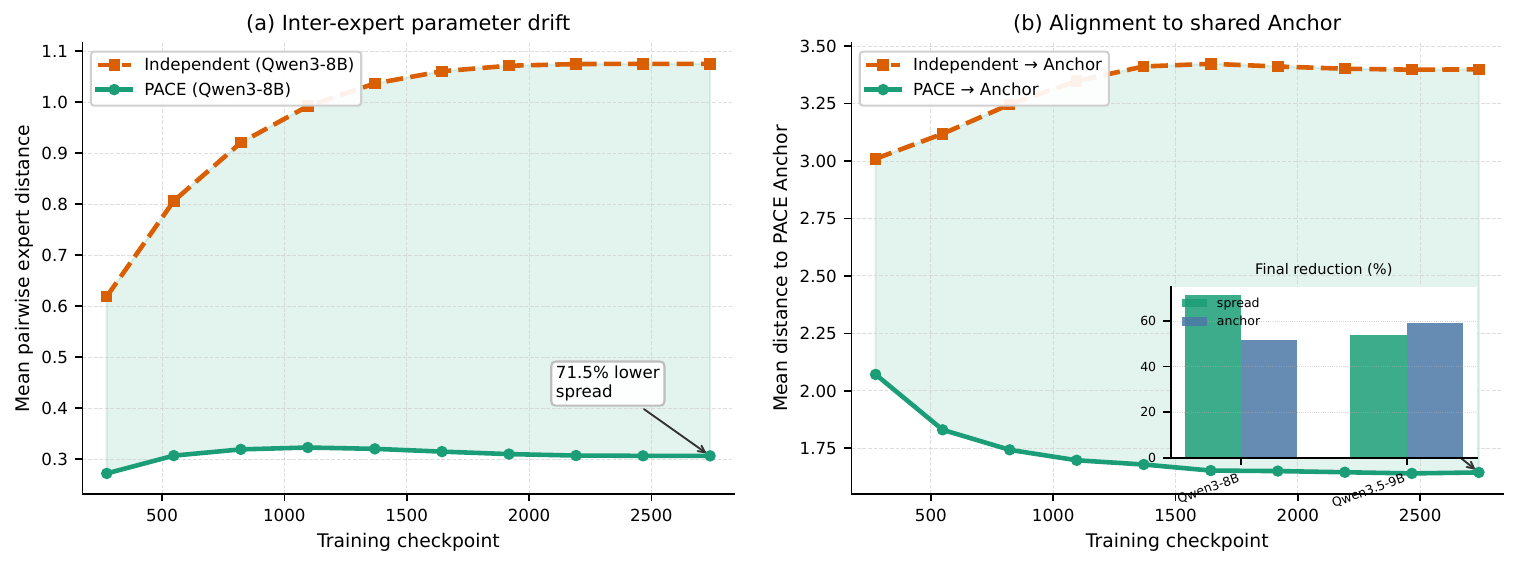}
    \caption{Parameter drift analysis across training checkpoints. (a) PACT substantially reduces the mean pairwise distance among paradigm-specific Branches compared with independent training. (b) PACT Branches remain closer to the shared Anchor, indicating that periodic aggregation and L1 Anchor Regularization align local paradigm updates to a common consensus. The inset reports final-checkpoint reductions on Qwen3-8B and Qwen3.5-9B.}
    \label{fig:pca_comparison}
\end{figure}

We measure Branch drift directly in LoRA parameter space rather than relying only on qualitative PCA projections. Under independent training, the mean pairwise Branch distance grows from 0.619 to 1.075 on Qwen3-8B, a 73.7\% increase. In contrast, PACT keeps the corresponding spread nearly flat, ending at 0.307. At the final checkpoint, PACT reduces inter-Branch spread by 71.5\% and Branch--Anchor distance by 51.6\% compared with independent training. The same trend appears on Qwen3.5-9B, where PACT reduces final Branch spread and Anchor distance by 54.0\% and 59.1\%, respectively.

\subsection{Sign Conflict Analysis}
\label{sec:sign_conflict}

To directly quantify information loss during TIES merging, we simulate the Trim$\to$Elect Sign pipeline offline at each checkpoint: we expand each Branch's LoRA delta $\Delta^{(k)} = B^{(k)} A^{(k)}$, apply magnitude trimming at density $\rho{=}0.5$, and measure the fraction of surviving positions where at least one Branch's sign disagrees with the elected majority (\textbf{sign conflict ratio}). We also report the \textbf{mean flip rate}: the average fraction of each Branch's active positions overridden by the majority sign.

Under independent training, sign conflict rises from 19.8\% to 38.0\% and saturates, confirming that Branches develop structurally incompatible parameter updates. The corresponding mean pairwise cosine between Branch task vectors drops from 0.61 to 0.18 (Figure~\ref{fig:teaser}). PACT reduces both metrics by over 99\%: final sign conflict falls to 0.38\% and mean flip rate to 0.11\%, with Branch cosine maintained at $\approx$1.0. This explains why PACT's periodic consensus yields higher merged-model quality (Table~\ref{tab:main}): near-zero conflict means TIES merging preserves virtually all paradigm-specific information.

\subsection{English-Translated Training and Evaluation}

\begin{table}[!t]
    \centering
    \resizebox{\columnwidth}{!}{
    \begin{tabular}{lcccc}
        \toprule
        Setting & Strict DA & Exam F1 & Avg Rnd & Symp Cov \\
        \midrule
        English Base & 33.39 & 20.13 & 3.73 & \textbf{77.05} \\
        English PACT & \textbf{47.83} & \textbf{38.78} & \textbf{1.27} & 62.79 \\
        \bottomrule
    \end{tabular}
    }
    \caption{English-translated dynamic evaluation on Qwen3-8B. DPS training data and held-out evaluation records are translated into English.}
    \label{tab:english_translation}
\end{table}

Table~\ref{tab:english_translation} evaluates whether PACT transfers under an English-translated diagnostic setting. Compared with the English Base model, English PACT improves Strict DA by +12.45 points, Soft Score by +7.78 points, and Exam F1 by +18.65 points, while reducing the average dialogue length from 3.73 to 1.27 rounds.

\end{document}